\documentclass{ceurart}

\usepackage{listings}
\usepackage{multirow}
\usepackage{graphicx}
\usepackage{siunitx} 
\usepackage{tabularx}
\usepackage{caption}
\usepackage{subcaption}
\begin{document}

\copyrightyear{2025}
\copyrightclause{Copyright for this paper by its authors.
  Use permitted under Creative Commons License Attribution 4.0
  International (CC BY 4.0).}

\conference{ESWC 2025: 2nd Workshop on Evaluation of Language Models in Knowledge Engineering (ELMKE),
  June 01--05, 2025, Portoroz, Slovenia}

\title{How do Scaling Laws Apply to Knowledge Graph Engineering Tasks? The Impact of Model Size on Large Language Model Performance}

\author[1,2]{Desiree Heim}[%
orcid=0000-0003-4486-3046,
email=desiree.heim@dfki.de,
url=,
]
\cormark[1]

\address[1]{DFKI, Kaiserslautern, Germany}
\address[2]{RPTU, Kaiserslautern, Germany}

\author[3,4]{Lars-Peter Meyer}[%
orcid=0000-0001-5260-5181,
email=lpmeyer@infai.org,
url=,
]
\address[3]{InfAI, Leipzig, Germany}
\address[4]{TU Chemnitz, Germany}
\author[1]{Markus Schröder}[%
orcid=0000-0001-8416-0535,
email=markus.schroeder@dfki.de,
url=,
]
\author[3,5]{Johannes Frey}[%
orcid=0000-0003-3127-0815,
email=frey@informatik.uni-leipzig.de,
url=,
]
\address[5]{Uni Leipzig, Germany}
\author[1,2]{Andreas Dengel}[%
orcid=0000-0002-6100-8255,
email=andreas.dengel@dfki.de,
url=,
]
\cortext[1]{Corresponding author.}

{\color{gray}\begin{picture}(0,0)
  \put(\dimexpr\paperwidth-1.6in\relax, -4.5in){%
    \makebox(0,0)[r]{\rotatebox{90}{\LARGE peer reviewed and to appear in the ESWC 2025 Workshops and Tutorials Joint Proceedings}}%
  }
\end{picture}
}

\begin{abstract}
When using Large Language Models (LLMs) to support Knowledge Graph Engineering (KGE), one of the first indications when searching for an appropriate model is its size. 
According to the scaling laws, larger models typically show higher capabilities. 
However, in practice, resource costs are also an important factor and thus it makes sense to consider the ratio between model performance and costs.
The LLM-KG-Bench framework enables the comparison of LLMs in the context of KGE tasks and assesses their capabilities of understanding and producing KGs and KG queries.
Based on a dataset created in an LLM-KG-Bench run covering 26 open state-of-the-art LLMs, we explore the model size scaling laws specific to KGE tasks.
In our analyses, we assess how benchmark scores evolve between different model size categories. 
Additionally, we inspect how the general score development of single models and families of models correlates to their size.
Our analyses revealed that, with a few exceptions, the model size scaling laws generally also apply to the selected KGE tasks.  
However, in some cases, plateau or ceiling effects occurred, i.e., the task performance did not change much between a model and the next larger model. 
In these cases, smaller models could be considered to achieve high cost-effectiveness.
Regarding models of the same family, sometimes larger models performed worse than smaller models of the same family. 
These effects occurred only locally. 
Hence it is advisable to additionally test the next smallest and largest model of the same family.
\end{abstract}


\maketitle

\section{Introduction}

Knowledge Graphs (KGs)~\cite{Hogan2020KnowledgeG} store facts about real-world domains in a structured way that facilitates reasoning to derive new information based on rules and existing knowledge. 
However, their creation and maintenance, commonly known as Knowledge Graph Engineering (KGE), usually requires manual, labour-intensive efforts including activities such as drafting an appropriate ontology, transforming data sources to fit the required format, and solving data integrity problems.
With the emergence of Large Language Models (LLMs), various approaches were developed with LLMs to support KGE tasks~\cite{PanUnifyingLLMsAndKGs,Allen2023KnowledgeEngineeringUsing,Buchmann2024Largelanguagemodels,Taffa2023LeveragingLLMsScholarly,hofer2022towards,Kovriguina2023SPARQLGENOP,Giglou2023LLMs4OLLargeLanguage}. 
Once LLMs are employed, the question arises of how well they can cope with KGs and KGE challenges.
To address this, the LLM-KG-Bench benchmark framework~\cite{Meyer2023DevelopingScalableBenchmark,Meyer2025LlmKgBench3} assesses the performance of LLMs on tasks requiring the comprehension of KGs~\cite{Frey2023Turtle,Frey2024AssessingEvolutionLLM}, their schemata and query languages~\cite{Meyer2025LlmKgBench3,Meyer2024AssessingSparqlCapabilititesLLM}.

The results on this benchmark not only show single model performances but might also provide valuable indications on KGE-specific scaling laws of LLMs.
Such scaling laws typically examine LLM task performances concerning their model sizes, training data size, or utilized computational training resources~\cite{Kaplan2020ScalingLF}.
Especially regarding model sizes, a usual expectation is that the larger the LLMs, the better their performances on downstream tasks.
Yet, this assumption can be wrong.
Moreover, larger models typically involve higher costs. 
In particular, the higher memory consumption of larger models compared to smaller ones is a highly relevant cost factor since more or more powerful hardware like GPUs are required.
At the same time, the parameter size also influences inference time assuming using the same hardware setting as with more parameters more weights need to be calculated.
Here, Mixture-of-Expert (MoE) models form an exception since only the number of active parameters, i.e., the proportion of parameters from the total parameters selected at runtime, influences the inference time.
Hence, when choosing between an MoE LLM and another one with the same amount of total parameters and a similar task performance, the MoE has a higher cost-effectiveness.
However, using the same hardware, for hosting either smaller and larger LLMs exclusively on, is, in practice, not necessarily realistic since smaller models would not fully exhaust the hardware's, e.g., GPUs, capability for the same setting, e.g., the same targeted number of concurrent request and maximum input lengths.
Hence, except for MoE models in comparison with similarly-sized LLMs, considering the memory requirement of LLMs is preferable.
In conclusion, considering the cost-effectiveness, i.e. particularly the memory requirements, the largest models may not be the best choice and a good trade-off between model performance and the model's resource demand has to be found.

In this paper, we therefore analyse LLM scaling laws on KGE tasks with respect to model sizes.
The data for our analysis are drawn from a recently published LLM-KG-Bench benchmark run~\cite{Meyer2025LlmKgBench3}.
It covers 26 open state-of-the-art LLMs from five providers and in total eleven model families, i.e., series of models released by a specific provider. 
Using the benchmark results and a combination of statistical analysis and visualizations, we would like to give initial answers to the following questions:
How do benchmark scores \dots
\begin{enumerate}
    \item \dots relate to different LLM model size groups?
    \item \dots develop with respect to model sizes in general?
    \item \dots develop with respect to model sizes within specific model families?
\end{enumerate}

By answering these questions, we aim to get more general insights about model size scaling laws on KGE tasks that can also be transferred to models not included in the benchmark run.

This paper is structured as follows: 
Section~\ref{related_work} introduces related works. 
In Section~\ref{experiment_setup}, we describe the LLM-KG Bench run and the obtained dataset used in our analyses. 
In Section~\ref{result_analysis}, we analyse the dataset in particular with respect to the correlation between model size and performance on the benchmark. 
Following the analysis, we summarize and discuss the gained insights in Section~\ref{discussion}. 
Section~\ref{conclusion} concludes this paper and gives an outlook to future work.

\section{Related Work}
\label{related_work}

In order to compare the vast amount of LLMs, there are several LLM leaderboards, which rank various LLMs based on a selection of benchmarks or workloads.
Among the well-known leaderboards are Chatbot Arena~\cite{chiang2024ChatbotArena}, which evaluates models using human preference on interactive tasks, and OpenLLM-Leaderboard~\cite{open-llm-leaderboard-v2}, covering numerous standard tasks like MMLU, BBH, and GPQU across more than $2,000$ models. 
Similarly, HELM~\cite{Liang2023HolisticEvaluationLanguageModels} provides comprehensive evaluations including domain-specific benchmarks such as LegalBench and MedQA. 

Regarding code generation, which shares a few conceptual similarities with KGE benchmarking, several dedicated benchmarks and leaderboards exist too.
Prominent code benchmarks include HumanEval and MultiPL-E, evaluated in the Big Code Models Leaderboard\footnote{\url{https://huggingface.co/spaces/bigcode/bigcode-models-leaderboard}}, as well as EvalPlus~\cite{liu2024your}, employing both HumanEval and the Mostly Basic Python Programming (MBPP) benchmark. 
The CanAiCode Leaderboard\footnote{\url{https://huggingface.co/spaces/mike-ravkine/can-ai-code-results}} specifically targets text-to-code tasks for smaller LLMs. 
These code-focused benchmarks emphasize structured output, syntactical correctness, and execution correctness, mirroring the evaluation criteria in KGE tasks, thereby offering insights relevant to benchmarking structured outputs from LLMs.

However, the mentioned attempts do not cover the evaluation of tasks specifically relevant to Knowledge Graph Engineering (KGE)~\cite{Meyer2023ExperimentsWithChatGPT}, such as RDF syntax correctness, SPARQL query generation, or graph comprehension. 

Efforts addressing KG-related evaluations frequently target specific problems like Text-to-RDF conversion~\cite{Mihindukulasooriya2023Text2KGBench,Zhu2023LlmsKgConstructionReasoning}, Knowledge Graph Question Answering (KGQA)~\cite{Usbeck2019Benchmarkingquestionanswering}, and SPARQL query generation~\cite{Kovriguina2023SPARQLGENOP,Zahera2024GeneratingSPARQLNatural}. 
These evaluations typically focus only on isolated tasks and often involve manual assessments, which limits scalability and adaptability to newer LLMs and task variations. 
An exception closely related to our interest in structured output is StructuredRAG~\cite{shorten2024structuredragjsonresponseformatting}, evaluating JSON-based structured responses from LLMs.

To address gaps in existing benchmarking efforts, especially regarding RDF and SPARQL tasks, LLM-KG-Bench~\cite{Meyer2023DevelopingScalableBenchmark,Meyer2025LlmKgBench3} provides a specialized automated benchmarking environment for evaluating semantic correctness and syntax handling in RDF and SPARQL tasks.
In contrast to general benchmarks like HELM or BigBench~\cite{srivastava2023imitationGameBigBench}, LLM-KG-Bench emphasizes semantic and syntactic correctness rather than multiple-choice accuracy, significantly reducing technological complexity for creating and evaluating KG-related tasks~\cite{Frey2023Turtle,Frey2024AssessingEvolutionLLM,Meyer2024AssessingSparqlCapabilititesLLM}.

Prior research already investigated the correlation between LLM parameter size and task performance~\cite{Kaplan2020ScalingLF,wei2022emergent}. 
Larger LLMs typically outperform smaller models for the same tasks but also exhibit emergent capabilities (not present in smaller models) such as complex reasoning or nuanced instruction-following abilities~\cite{wei2022emergent,hoffmann2022training}.
However, this relationship is not universally linear; task type, complexity, and input and output structure can significantly influence whether larger models yield proportionally better performance. 
Scenarios and tasks w.r.t. Knowledge Graph Engineering, which typically requires dealing with RDF serialization formats and paradigms, remains underexplored. 
This study addresses this gap by explicitly examining how model size impacts performance across diverse RDF and SPARQL tasks within the context of KG engineering.

\section{Dataset}
\label{experiment_setup}

This work analyses data generated by the LLM-KG-Bench framework~\cite{Meyer2023DevelopingScalableBenchmark,Meyer2025LlmKgBench3}.
The LLM-KG-Bench framework offers the infrastructure to define KG engineering-related automated tasks that can be repeatedly executed on many LLMs to evaluate their performance.
Since also the evaluation is automated, the same experiments can be repeated which increases reproducibility and gives a broader sample size for statistical analysis to take the probabilistic nature of LLM-generated answers into account.

The dataset used in this work evaluated over 30 open and proprietary LLMs on 26 RDF- and SPARQL-related task variations.

The dataset contains LLMs from three open LLM providers: Qwen, Meta-LLama, and Microsoft-Phi. They were selected because of providing official instruction-finetuned Large Language Models that were the highest-ranked on the Open LLM Leaderboard~\cite{open-llm-leaderboard-v2} in December 2024 with respect to their average score across all benchmarks included in the leaderboard~\footnote{Upstage providing the solar LLM family was excluded here since the models only support a maximum context length of up to 4096k Token which was not sufficient for all tasks}. We restricted our selection to models of up to 80B parameters due to restrictions on hardware resources available to us.
In addition to that, the dataset includes three LLMs fine-tuned or optimized for code understanding and generation which also requires handling structured data similar to KG-related tasks: Qwen2.5-Coder-Instruct-32B, Infly-OpenCoder-8B-Instruct, and deepseek-coder-33b-instruct. For the selection of these models, we consulted the Mostly Basic Python Programming (MBPP) Benchmark score reported by the EvalPlus Leaderboard \cite{liu2024your} and decided for the top-ranked instruction-finetuned models not larger than 80B parameters that are reported to be explicitly optimized or fine-tuned for code.

The model sizes range from $0.5$ billion parameters up to 72 billion parameters.
Two included LLMs are Mixture-of-Experts models: Qwen2-Instruct with 57 billion parameters (14 billion active) and Phi-3.5-instruct with 42 billion parameters ($6.6$ billion active).
With mixture-of-expert models only a subset of parameters is active during inference, resulting in a lower effective parameter count compared to the total model size.
An overview of the evaluated models can be found in Table~\ref{tab:ModelDetails}. 
More details on the models and their selection can be found in the dedicated paper~\cite{Meyer2025LlmKgBench3}.

In addition to the open LLMs, several proprietary LLMs from the OpenAI GPT, Google Gemini
and Anthropic Claude families that achieved constantly high scores on the  Chatbot Arena Leaderboard~\cite{chiang2024ChatbotArena} were included in the benchmark run, namely ChatGPT 3.5 turbo, ChatGPT 4o, ChatGPT 4o-mini, ChatGPT o1, ChatGPT o1-mini, Gemini 2.0 Flash, Gemini 1.5 Pro, Gemini 1.5 Flash, Claude 3.5 Sonnet and Claude 3.5 Haiku.
However, since model sizes for proprietary LLMs are not documented, we selected only the remaining 26 open LLMs for our main analysis and refer to the achieved scores of the proprietary models only briefly for comparison to better classify the open LLM performance.

From the 26 task variations included in the dataset, we analyse 23 variations of seven task classes in the KG engineering areas of RDF and SPARQL handling.
To focus the comparison on various input formats for consistency reasons, three task variations of the \textit{Text2Sparql} consisting of other KGs as datasets were excluded and eight task variations of \textit{RdfFriendCount} in the dataset were aggregated into four task variations for the analysis. 
Task variations of a task class have a similar prompt and evaluation but differ e.g. by the serialization format (JSON-LD, N-Triples, Turtle, XML) presented to the LLM.

For each open LLM, respectively 50 repetitions per task variation were executed.  
To assess the performance of the LLMs, tasks compute several measures based on the LLM answers with values in the interval $[0,1]$ with better answers resulting in higher scores.
These measures often include ones based on recall, precision, and f1~measure as well as e.g. brevity measures or ones indicating if the answer was at least syntactically correct.
For some tasks, there are variations of measures defined with different levels of strictness in the evaluation.

We selected measures that examine the result correctness in sufficiently different ways to provide a concise overview.
Therefore, brevity measures are not included and F1-based measures were selected over precision- and recall-based measures. 
In the case of similar measures, only one representative was chosen, e.g. measures that check the responses relying on the requested output format or measures that search the answer for expected components.
Here, stricter measures have been preferred to more relaxed ones. 
Regarding measures that operate on output lists, we selected measures that remove leading and trailing white space, since it is only a minor correction.
Additionally, for tasks yielding RDF graphs or SPARQL queries, measures indicating their syntactical correctness were included.

The different calculated measures can be classified into three types:
\begin{description}
    \item[Central] These format-sensitive answer quality measures assess the answer correctness sensitive to the instructed output format, i.e., the output accuracy is assessed assuming that the requested format is respected. They are \textit{listTrimF1}, \textit{f1}, \textit{strSimilarity} and \textit{trimF1}.
    \item[Fragment] The fragment-based answer quality measures measure the answer correctness but are less strict regarding the answer format when evaluating the answer and account for correct answer parts. They include \textit{textHttpF1}, \textit{contentF1} and \textit{sparqlIrisF1}.
    \item[Syntax] Syntactical answer correctness measures inspect whether the output is syntactically correct, i.e. fulfills all criteria for valid graphs or queries. The two measures \textit{parsableSyntax} and \textit{answerParse} belong to this category.
\end{description}

In the following, the seven task classes and the measures selected for this analysis are briefly described.
More information can be found in the LLM-KG-Bench documentation\footnote{Task documentation: \url{https://github.com/AKSW/LLM-KG-Bench/blob/v3.0.0/LlmKgBench/tasks/README.md}} or in the articles introducing them~\cite{Meyer2025LlmKgBench3,Frey2023Turtle,Frey2024AssessingEvolutionLLM,Meyer2024AssessingSparqlCapabilititesLLM}.

\begin{table}
    \caption{
       Model sizes for the 26 open LLMs analysed grouped by model family. 
    Two of these models utilize the mixture of experts architecture (denoted by *), where only a subset of parameters is active during inference. 
    }
    \label{tab:ModelDetails}
    \centering
    \begin{tabular}{lcccccccc}
    \toprule
Model (Family) Name                  & \multicolumn{8}{l}{Model Sizes = Number of Parameters}  \\
    \midrule
Qwen2-Instruct              & 0.5B  & 1.5B    &         & 7B &     &         & 57B*    & 72B    \\
Qwen2.5-Instruct            & 0.5B  & 1.5B    & 3B      & 7B & 14B & 32B     &         & 72B    \\
Qwen2.5-Coder-Instruct      &       &         &         &    &     & 32B     &         &        \\
    \cmidrule{1-1}
Meta-LLama-3-Instruct       &       &         &         & 8B &     &         &         & 70B    \\
Meta-LLama-3.1-Instruct     &       &         &         & 8B &     &         &         & 70B    \\
Meta-LLama-3.2-Instruct     &       & 1B      & 3B      &    &     &         &         & 70B    \\
Meta-LLama-3.3-Instruct     &       &         &         &    &     &         &         & 70B    \\
    \cmidrule{1-1}
Microsoft-Phi-3-instruct    &       &         & 3.8B    & 7B & 14B &         &         &        \\
Microsoft-Phi-3.5-instruct  &       &         & 3.8B    &    &     &         & 42B*    &        \\
    \cmidrule{1-1}
Infly-OpenCoder-8B-Instruct &       &         &         & 8B &     &         &         &        \\
    \cmidrule{1-1}
deepseek-coder-33b-instruct &       &         &         &    &     & 33B     &         &        \\
    \bottomrule
     \end{tabular}
\end{table}

\paragraph{RdfConnectionExplain} This task consists of finding the shortest connection between two nodes in a small KG which requires a basic understanding of serialization formats and RDF concepts.
There are four variations of this task. Each presents the graph in a different serialization format: JSON-LD, N-Triples, Turtle, or RDF/XML.
Here, a list of IRIs representing the shortest path is the expected answer format.
For the given answer, the task computes \textit{listTrimF1} as F1-measure on trimmed list entries without leading or trailing whitespaces.
The \textit{textHttpF1} measure is an F1-measure on IRI-like answer parts starting, e.g., with ``\verb|http://|''.

\paragraph{RdfFriendCount} This task presents a small KG with nodes of one type and edges of one type.
The LLM is asked to return the node with the most incoming edges.
There are 4 KG serialization format variations: JSON-LD, N-Triples, Turtle, and RDF/XML.
The task computes the \textit{f1} measure on the nodes found in the answer.

\paragraph{RdfSyntaxFixing} A KG with syntax errors is provided and the LLM is queried to correct it. There are 3 variations introduced with the serialization formats JSON-LD, N-Triples, and Turtle.
The measure \textit{parsableSyntax} equals $1$ if the RDF syntax in the answer is parsable ($0$ otherwise).
\textit{strSimilarity} is computed by comparing the given RDF with the expected answer, and \textit{contentF1} is the F1-measure on the given RDF graph on a triple level.

\paragraph{Sparql2Answer} In this task, the LLM is asked to respond with the result set for a given SPARQL SELECT query given the KG.
There are 2 variations with the graph serialization formats JSON-LD and Turtle.
The answer should be a list of entities and the \textit{trimF1} measure is computed as F1-measure on the trimmed entities, where leading and trailing whitespaces are removed.

\paragraph{SparqlSyntaxFixing} Similar to the RdfSyntaxFixing task, the LLM is asked to fix syntactically erroneous SPARQL SELECT queries.
The measure \textit{answerParse} equals $1$ if the adapted SPARQL query syntax is correct ($0$ otherwise).
\textit{sparqlIrisF1measure} is the F1-measure on the IRIs found in the modified SPARQL query.
\textit{f1measure} refers to the result set obtained when executing the corrected SPARQL SELECT query.

\paragraph{Text2Answer} Similar to the Sparql2Answer task, the LLM is asked to respond with the result set for a given natural language question given a small KG.
There are 2 variations of the graph presented in the serialization formats JSON-LD and Turtle.
Similar to the \textit{Sparql2Answer} task, the answer is expected as a list and the \textit{trimF1} measure is computed on the trimmed list elements.

\paragraph{Text2Sparql} Here a natural language question is presented together with information on a KG and the LLM is asked to translate the question into a suitable SPARQL SELECT query.
There are 3 variations with the KG presented in the form of a complete schema, only the relevant schema or the relevant subgraph, all in Turtle syntax.
The same measures as described for the \textit{SparqlSyntaxFixing} task were selected:
\textit{answerParse}, \textit{sparqlIrisF1measure} and \textit{f1measure}.
\newline
For all tasks, the prompts are kept relatively simple and are not specifically optimized using prompt engineering to assess the basic capabilities of LLMs.
Moreover, we restrict from using advanced prompting techniques to prevent certain models from gaining an unfair advantage that could occur after they are used in the prompt engineering process.
In the following section, we analyse the described dataset.

\section{Result Analysis}
\label{result_analysis}

In this section, we report and analyse the results of the benchmark run.
First, the overall task performance is examined to explore task-centered tendencies (Section~\ref{sec41}).
Second, we take a closer look at task performances with respect to model sizes and shed light on two aspects: comparison of performances between different size categories (Section~\ref{sec42}) and the development of scores with respect to model sizes and families (Section~\ref{sec43}).

\subsection{Overall Task Performance}
\label{sec41}

To get an overview of the benchmark scores achieved by the open LLMs included in the experiments, Table~\ref{tab:task_view_benchmark} lists the means and standard deviations of all LLM scores per task variation.
Additionally, mean scores of individual LLMs as well as the highest and lowest intra-LLM mean are reported.

Regarding mean calculation, missing values of central and fragment measures originating from unparsable RDF or SPARQL outputs were filled with 0 to account for the fact that the outputs do not even meet the minimum quality criterion of syntactical correctness.
Concerning tasks that allow corrections to initial answers (i.e. multiple Prompt-Answer-Evaluate loops), only the last answer scores are considered in the table, since the mean of all scores for the first and last answers show only minor differences.

In the following, we will examine the scores for each measure type that are listed in Table \ref{tab:task_view_benchmark}.
\begin{table}[t]
\caption{This table depicts data about the selected benchmark scores grouped by tasks and task variations. It presents basic statistics including mean, standard deviation (std), minimum (min) and maximum (max) mean achieved by one specific LLM. For each measure, the respective type is provided. In case a task allows for up to two retries, it is marked with (*) and the scores achieved by the last output are given.}\label{tab:task_view_benchmark}
\begin{tabularx}{\textwidth}{LLLLRRRR}
\toprule
  Task &
  Variation &
  Measure &
  Type &
  mean &
  std &
  min &
  max \\
\midrule
RdfConnectionExplain &
  jsonld &
  listTrimF1 &
  central &
  0.64 &
  0.36 &
  0.03 &
  1.00 \\
 &
   &
  textHttpF1 &
  fragment &
  0.71 &
  0.30 &
  0.03 &
  1.00 \\ \cmidrule(r){2-3}
 &
  nt &
  listTrimF1 &
  central &
  0.51 &
  0.34 &
  0.02 &
  0.95 \\
 &
   &
  textHttpF1 &
  fragment &
  0.68 &
  0.25 &
  0.34 &
  0.98 \\ \cmidrule(r){2-3} 
 &
  turtle &
  listTrimF1 &
  central &
  0.68 &
  0.32 &
  0.08 &
  1.00 \\
 &
   &
  textHttpF1 &
  fragment &
  0.75 &
  0.27 &
  0.30 &
  1.00 \\ \cmidrule(r){2-3} 
 &
  xml &
  listTrimF1 &
  central &
  0.70 &
  0.30 &
  0.12 &
  1.00 \\ 
 &
   &
  textHttpF1 &
  fragment &
  0.77 &
  0.23 &
  0.42 &
  1.00 \\ \cmidrule(r){1-3} 
RdfFriendCount &
  jsonld &
  f1 &
  central &
  0.17 &
  0.37 &
  0.00 &
  1.00 \\ 
 &
  nt &
  f1 &
  central &
  0.17 &
  0.37 &
  0.00 &
  1.00 \\  
 &
  turtle &
  f1 &
  central &
  0.06 &
  0.22 &
  0.00 &
  0.47 \\  
 &
  xml &
  f1 &
  central &
  0.29 &
  0.45 &
  0.00 &
  1.00 \\ \cmidrule(r){1-3}
RdfSyntaxFixing * &
  jsonld &
  parsableSyntax &
  syntax &
  0.81 &
  0.39 &
  0.00 &
  1.00 \\ 
 &
   &
  strSimilarity &
  central &
  0.67 &
  0.38 &
  0.10 &
  0.87 \\  
 &
   &
  contentF1 &
  fragment &
  0.78 &
  0.40 &
  0.00 &
  1.00 \\ \cmidrule(r){2-3} 
 &
  nt &
  parsableSyntax &
  syntax &
  0.74 &
  0.44 &
  0.00 &
  1.00 \\ 
 &
   &
  strSimilarity &
  central &
  0.61 &
  0.43 &
  0.08 &
  1.00 \\ 
 &
   &
  contentF1 &
  fragment &
  0.65 &
  0.46 &
  0.00 &
  1.00 \\ \cmidrule(r){2-3} 
 &
  turtle &
  parsableSyntax &
  syntax &
  0.68 &
  0.47 &
  0.02 &
  1.00 \\  
 &
   &
  strSimilarity &
  central &
  0.42 &
  0.34 &
  0.10 &
  0.90 \\ 
 &
   &
  contentF1 &
  fragment &
  0.67 &
  0.46 &
  0.01 &
  1.00 \\ \cmidrule(r){1-3}
Sparql2Answer &
  jsonld &
  trimF1 &
  central &
  0.54 &
  0.47 &
  0.01 &
  1.00 \\ 
 &
  turtle &
  trimF1 &
  central &
  0.58 &
  0.47 &
  0.01 &
  1.00 \\ \cmidrule(r){1-3}
SparqlSyntaxFixing * &
   &
  answerParse &
  syntax &
  0.68 &
  0.47 &
  0.00 &
  1.00 \\ 
 &
   &
  f1measure &
  central &
  0.60 &
  0.49 &
  0.00 &
  1.00 \\ 
 &
   &
  sparqlIrisF1measure &
  fragment &
  0.66 &
  0.46 &
  0.00 &
  1.00 \\ \cmidrule(r){1-3}
Text2Answer &
  jsonld &
  trimF1 &
  central &
  0.57 &
  0.48 &
  0.02 &
  1.00 \\ 
 &
  turtle &
  trimF1 &
  central &
  0.63 &
  0.47 &
  0.03 &
  1.00 \\ \cmidrule(r){1-3}
Text2Sparql * &
  turtle schema &
  answerParse &
  syntax &
  0.72 &
  0.45 &
  0.00 &
  1.00 \\  
 &
   &
  f1measure &
  central &
  0.13 &
  0.27 &
  0.00 &
  0.31 \\ 
 &
   &
  sparqlIrisF1measure &
  fragment &
  0.30 &
  0.30 &
  0.00 &
  0.53 \\ \cmidrule(r){2-3}
 &
  turtle subschema &
  answerParse &
  syntax &
  0.74 &
  0.44 &
  0.06 &
  1.00 \\ 
 &
   &
  f1measure &
  central &
  0.10 &
  0.25 &
  0.00 &
  0.28 \\ 
 &
   &
  sparqlIrisF1measure &
  fragment &
  0.35 &
  0.29 &
  0.01 &
  0.57 \\ \cmidrule(r){2-3}
 &
  turtle subgraph &
  answerParse &
  syntax &
  0.81 &
  0.39 &
  0.06 &
  1.00 \\ 
 &
   &
  f1measure &
  central &
  0.57 &
  0.45 &
  0.00 &
  0.93 \\
 &
   &
  sparqlIrisF1measure &
  fragment &
  0.71 &
  0.40 &
  0.04 &
  0.96 \\ 
  \bottomrule
\end{tabularx}
\end{table}

Regarding the mean scores of \textbf{central measures} they are in average medium-high being close to a score of $0.6$ on the tasks \textit{SparqlSyntaxFixing}, \textit{RdfConnectionExplain}, \textit{Text2Answer}, \textit{RdfSyntaxFixing} and \textit{Sparql2Answer}. 
In contrast, \textit{RdfFriendCount} stands out with low means between $0.06$ and $0.29$. 
For the \textit{Text2SPARQL} task, the two input variations turtle schema and subschema also got low scores of $0.13$ and $0.10$, while the input variation turtle subgraph achieved a comparably high mean score of $0.57$. 
For other tasks, the difference between the mean scores for input variations is comparably small with a maximum difference of $0.25$ between the lowest and highest mean in the task class \textit{RdfSyntaxFixing}. 
Moreover, no clear task-overarching preference for a specific KG format (turtle, nt, jsonld, xml) is recognizable.

Looking at the standard deviation, the scores on the central measures are widespread. 
Roughly 20\% of the central measures have a standard deviation between $0.2$ and $0.3$, 40\% have one larger than $0.3$ to $0.4$ and the remaining 40\% have a dispersion larger than $0.4$ to $0.5$. 
This is also reflected in the minimum and maximum average central measure score per LLM. 
The highest minimum mean is $0.12$ while, except for three outliers, the maximum means are $0.75$ or higher, and most have even a mean of 1 or close values. 
Here, only the turtle input variation of the \textit{RdfFriendCount} task with a maximum score of $0.47$ and the turtle schema and turtle subschema input variations of the \textit{Text2SPARQL} task with a maximum intra-LLM mean of $0.31$ and $0.28$ differ substantially from the other means. 
In all of these cases, this circumstance is also apparent in a low overall mean score.

The \textbf{fragment measures} show tendencies similar to the central measures. 
As expected, for all task variations their means are higher compared to the central measures.
This is also reflected in the minimum and maximum intra-LLM means. 
Only for the \textit{RdfSyntaxFixing} task, the minimum mean per LLM is lower than those of the central measures. 
Notably, in contrast to the central measures, the fragment measures are only calculated if the output graph is syntactically correct, otherwise, they are 0.

Last but not least, the means of \textbf{syntax measures} for the \textit{RdfSyntaxFixing}, \textit{SparqlSyntaxFixing}, and \textit{Text2Sparql} tasks are rather high ranging between $0.68$ and $0.81$. 
However, the dispersion of values around the means is relatively high with standard deviations between $0.39$ and $0.47$. 
Without exceptions, the minimum intra-LLM means are all close or equal to $0$, and the maximums close or equal to $1$.

This subsection gave an overview of the models' performances on all tasks to classify task classes and their variations. 
Building upon that, the following two subsections focus on comparing the models' answer quality based on their sizes, i.e., trained parameters.
First, in Section \ref{sec42}, task performances are compared between model size categories, and in Section \ref{sec43} the development of scores in general and within families is visually assessed with respect to model sizes.

\subsection{Size Category Performance Similarities}
\label{sec42}

In the following analyses, only the central measures are included since they indicate most accurately whether a given answer is correct also taking into account the adherence to the requested output format.
To examine whether LLM size affects task performance, we first divided the LLMs into four groups with respect to their sizes. 
We classify models into the size categories tiny $[0-3B]$, small $(3B-8B]$, medium $(8B-33B]$ and large $(33B-72B]$.
Subsequently, to assess whether there are any significant differences in achieved central benchmark scores between the LLM size groups, we conducted Kruskal–Wallis tests \cite{Kruskal1952UseOR} for each task variation with the null hypothesis that the score distributions of all groups are identical.  
For all tests performed, null hypotheses were rejected with a significance level of less than $0.001$ indicating that for all task variations, there are significant differences between model size groups.
The highest significance level was obtained for the \textit{RdfConnectionExplain xml} variation with p$\approx$$5e-12$ and the lowest significance level was p$\approx$$7e-122$ for the \textit{Text2Sparql turtle subschema} variation indicating highly significant differences between the groups.

Since the Kruskal-Wallis test only measures whether there is a significant difference between a set of groups, next, posthoc Dunn tests \cite{Dunn1964MultipleCU} with Bonferroni correction \cite{Bonferroni1935IlCD} were conducted to examine which groups are dissimilar. 
Again, the null hypothesis was that there is no difference between the group pairs.
Table \ref{tab:dunnbonferroni} shows the results of the post-hoc tests for each task variation. 
Group pairs that are dissimilar with a significance of 5\% or less are blank. 
For all pairs not classified as dissimilar, the p-value is provided.
The higher this value is, the more similar groups can be considered.
Additionally, on the left-hand side of the table, the mean scores per model size group are given as a reference.
Groups with a high standard deviation $(0.3, 0.4]$ were marked with a $\sim$, and those with a very high standard deviation $(0.4, 0.5]$ were marked with $\approx$.
All other groups have a standard deviation of $0.3$ or lower.

Overall, as expected, most comparisons show, with a significance of 5\%, a dissimilarity between the respective group pairs, i.e., their respective scores were significantly different (the null hypotheses were rejected). 
Except for six dissimilar pairs, the differences were also very significant with $p<0.001$.
Typically, the identified significant score differences between groups are associated with rising scores from groups containing smaller LLMs to groups with larger model sizes.
However, only the input variations turtle and xml of the \textit{RdfFriendCount} Task show unexpectedly decreasing scores from a group of smaller to a group of larger LLMs.
Additionally, deviating from the assumptions, there were also pairs for which no significant differences were recognizable.
For these groups, the p-value is given in Table \ref{tab:dunnbonferroni}.
Higher values indicate that size categories can be considered more similar with respect to their task performance.
For the pairs of medium and large groups, this applies to roughly half of the cases.
Mostly, for both groups, the average scores were high, i.e., a ceiling effect occurred. 
In three cases, for the \textit{RdfFriendCount nt}, \textit{Text2Sparql turtle schema} and \textit{Text2Sparql turtle subschema} task variations the overall scores are low and show plateau effects, i.e. they do not change perceivably, so no significant differences were detected.
The pair with the second highest number of insignificant score differences is small-medium with five cases.
Here, mainly plateau effects occurred.
Similarly, the three insignificant differences between scores of the pair tiny-small are plateaus. 
In contrast to groups adjacent with respect to the size class they represent, pairs that are not directly adjacent have predominately significantly different scores. 
Hence, there are only two cases of rather similar scores forming plateaus between the groups tiny and medium and one case between the pair small and large.

In summary, overall the most frequent perceivable effect is a rise of the average scores from smaller to larger model groups. 
Notable exceptions are the task variations turtle and xml \textit{RdfFriendCount}, for which the scores between smaller and larger model size groups decrease significantly.
Other than that, plateaus occur for which the scores between adjacent groups do not change significantly.
However, in some cases, the plateaus occur only locally and a rise of scores is detectable in particular between the medium and large groups (see e.g. \textit{RdfFriendCount jsonld} or \textit{RdfConnectionExplain nt}).
For some task variations, the scores of groups medium and large also almost reach the upper score bound of 1.

\begin{table}
\caption{
  The table shows the mean scores of central measures per model size groups and the group-wise similarity.
  The models are grouped into the size groups tiny (t, $[0-3B]$), small (s, $(3B-8B]$), medium (m, $(8B-33B]$) and large (l, $(33B-72B]$).
  Mean scores with a high standard deviation (0.3 to 0.4) were marked with a $\sim$ and those with a very high standard deviation (0.4 to 0.5) were marked with $\approx$.
  For the group-wise similarity only p-values $>$0.05 of post-hoc test are shown. In the other cases, groups show statistically significant differences.
  Higher p-values indicate more similar groups.
}
\resizebox{\textwidth}{!}{
\begin{tabular}{llSSSSSSSSSS}
\toprule
& &
  \multicolumn{4}{c}{mean score per model size group} &
  \multicolumn{6}{c}{group-wise similarity} \\ 
  \cmidrule(lr){3-6}
  \cmidrule(l){7-12}
  Task  &
  Variation &
  {t} &
  {s} &
  {m} &
  {l} &
  {t$\leftrightarrow$ s} &
  {t$\leftrightarrow$ m} &
  {t$\leftrightarrow$ l} &
  {s$\leftrightarrow$ m} &
  {s$\leftrightarrow$ l} &
  {m$\leftrightarrow$ l} \\ 
  \midrule
 \multirow{4}{*}{\begin{tabular}[c]{@{}c@{}}RdfConnection\\Explain\end{tabular}} &
  jsonld &
  {$\sim$} 0.41 &
  0.61 &
  {$\sim$} 0.75 &
  0.87 &
   &
   &
   &
   &
   &
   \\  
 &
  nt &
  {$\sim$} 0.39 &
  0.51 &
  0.49 &
  {$\sim$} 0.68 &
   &
  0.07 & 
   &
  1.00 & 
   &
  \\  
 &
  turtle &
  {$\sim$} 0.50 &
  0.59 &
  0.82 &
  0.87 &
   &
   &
   &
   &
   &
  .13 
   \\ 
 &
  xml &
  0.53 &
  0.60 &
  0.79 &
  0.93 &
  0.09 & 
   &
   &
   &
   &
   \\ 
  \cmidrule{1-2}
 \multirow{4}{*}{\begin{tabular}[c]{@{}c@{}}RdfFriend\\ Count\end{tabular}} &
  jsonld &
  0.09 &
  0.09 &
  0.06 &
  {$\approx$} 0.41 &
  1.00 &
  1.00 & 
   &
  1.00 & 
   &
   \\ 
 &
  nt &
  0.04 &
  {$\sim$} 0.13 &
  {$\approx$} 0.26 &
  {$\approx$} 0.32 &
   &
   &
   &
   &
   &
  1.00 \\ 
 &
  turtle &
  0.05 &
  0.03 &
  {$\sim$} 0.18 &
  0.01 &
  1.00 & 
   &
   &
   &
   &
   \\ 
 &
  xml &
  {$\approx$} 0.31 &
  {$\sim$} 0.11 &
  0.07 &
  {$\approx$} 0.57 &
   &
   &
   &
  1.00 & 
   &
   \\ 
  \cmidrule{1-2}
\multirow{3}{*}{\begin{tabular}[c]{@{}c@{}}RdfSyntax\\ Fixing\end{tabular}} &
  jsonld &
  {$\approx$} 0.59 &
  {$\approx$} 0.68 &
  0.98 &
  0.96 &
   &
   &
   &
   &
   &
  1.00 \\ 
 &
  nt &
  {$\approx$} 0.36 &
  {$\approx$} 0.61 &
  {$\sim$} 0.86 &
  0.91 &
   &
   &
   &
   &
   &
  0.60 \\ 
 &
  turtle &
  {$\approx$} 0.34 &
  {$\approx$} 0.66 &
  0.95 &
  0.92 &
   &
   &
   &
   &
   &
  1.00 \\ 
  \cmidrule{1-2}
\multirow{2}{*}{\begin{tabular}[c]{@{}c@{}}Sparql2Answer\end{tabular}} &
  jsonld &
  {$\sim$} 0.25 &
  {$\approx$} 0.42 &
  {$\approx$} 0.75 &
  {$\sim$} 0.88 &
   &
   &
   &
   &
   &
   \\
 &
  turtle &
  {$\sim$} 0.24 &
  {$\approx$} 0.54 &
  {$\sim$} 0.76 &
  0.92 &
   &
   &
   &
   &
   &
   \\ 
  \cmidrule{1-2}
\begin{tabular}[c]{@{}c@{}}SparqlSyntax\\ Fixing\end{tabular} &
   &
  {$\approx$} 0.25 &
  {$\approx$} 0.50 &
  {$\sim$} 0.86 &
  0.95 &
   &
   &
   &
   &
   &
  0.12 \\ 
  \cmidrule{1-2} 
\multirow{2}{*}{\begin{tabular}[c]{@{}c@{}}Text2Answer\end{tabular}} &
  jsonld &
  {$\approx$} 0.26 &
  {$\approx$} 0.50 &
  {$\approx$} 0.73 &
  0.92 &
   &
   &
   &
   &
   &
   \\ 
 &
  turtle &
  {$\approx$} 0.28 &
  {$\approx$} 0.64 &
  {$\approx$} 0.76 &
  0.97 &
   &
   &
   &
   &
   &
   \\ \cmidrule{1-2}
\multirow{3}{*}{\begin{tabular}[c]{@{}c@{}}Text2Sparql\end{tabular}} &
  schema &
  0.04 &
  0.13 &
  0.15 &
  {$\sim$} 0.23 &
   &
   &
   &
  1.00 & 
   &
  0.21 \\ 
 &
  subschema &
  {$\sim$} 0.04 &
  {$\approx$} 0.12 &
  0.16 &
  0.13 &
   &
   &
   &
  1.00 & 
  1.00 & 
  0.34 \\
 &
  subgraph &
  0.19 &
  0.58 &
  0.84 &
  0.86 &
   &
   &
   &
   &
   &
  1.00 \\ 
\bottomrule
\end{tabular}
}
\label{tab:dunnbonferroni}
\end{table}

\subsection{Task Performance by Model Size and Family}
\label{sec43}

\begin{figure} 
    \centering
    \begin{subfigure}{0.33\textwidth}
        \centering
        \includegraphics[width=\linewidth]{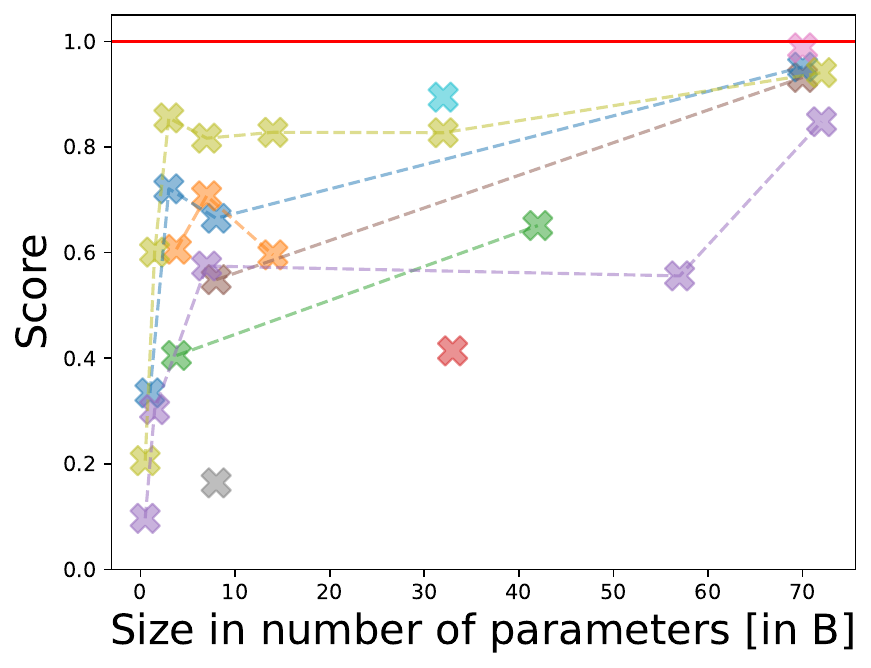} 
        \caption{RdfConnectionExplain}
    \end{subfigure}\hfill
    \begin{subfigure}{0.33\textwidth}
        \centering
        \includegraphics[width=\linewidth]{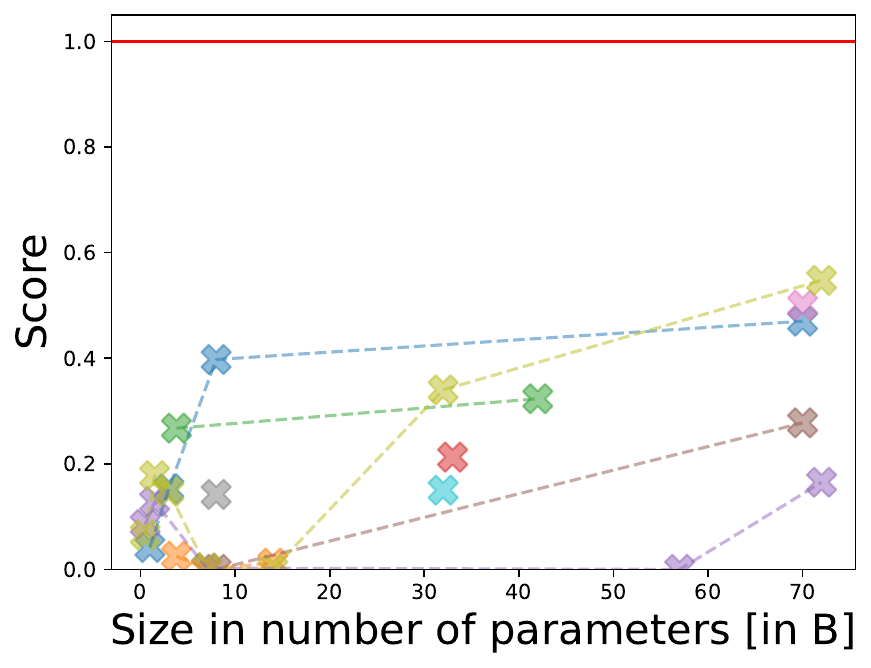}
        \caption{RdfFriendCount}
    \end{subfigure}\hfill
    \begin{subfigure}{0.33\textwidth}
        \centering
        \includegraphics[width=\linewidth]{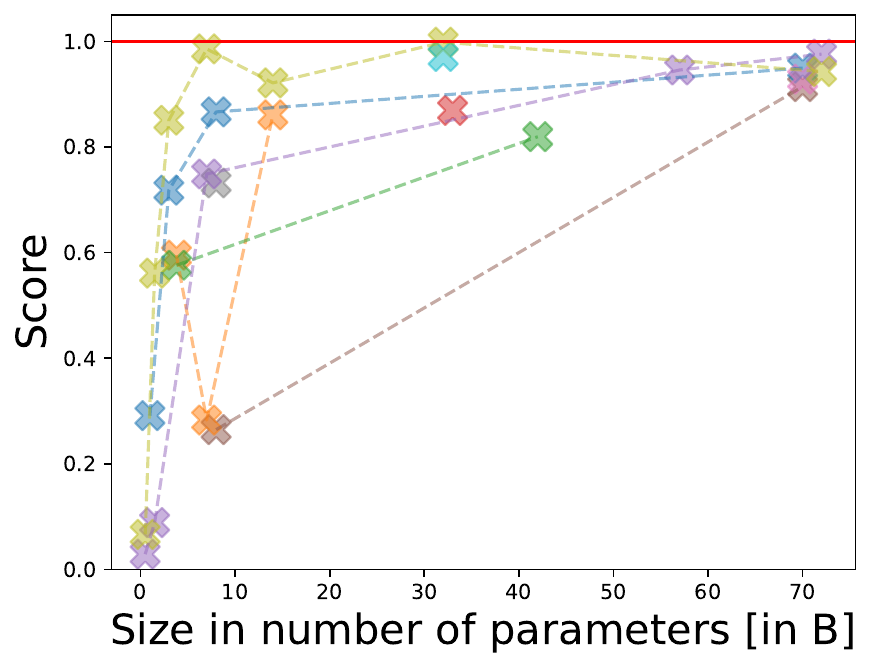}
        \caption{RdfSyntaxFixing}
    \end{subfigure}
    
    \vskip 0.3cm 
    
    \begin{subfigure}{0.33\textwidth}
        \centering
        \includegraphics[width=\linewidth]{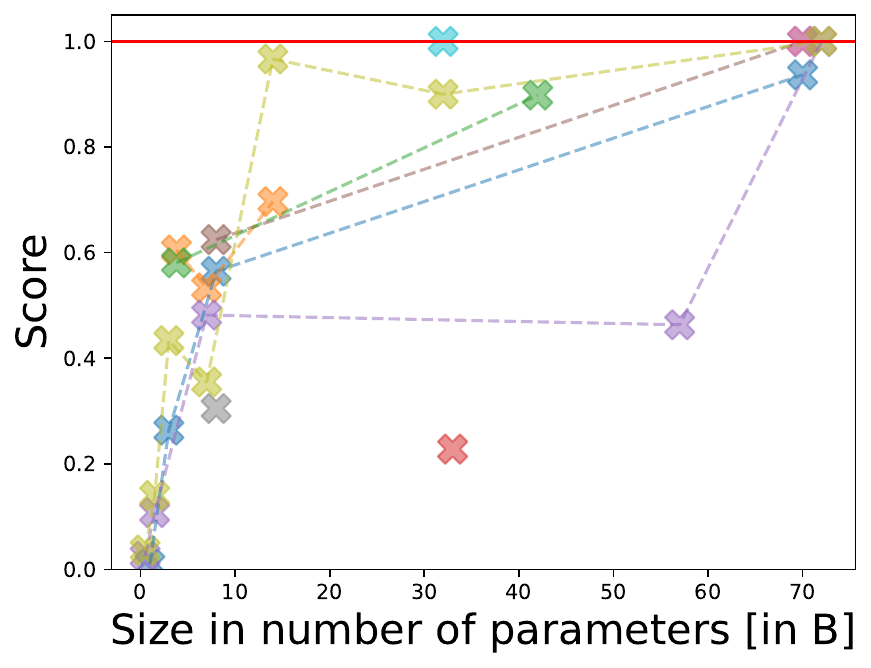}
        \caption{Sparql2Answer}
    \end{subfigure}\hfill
    \begin{subfigure}{0.33\textwidth}
        \centering
        \includegraphics[width=\linewidth]{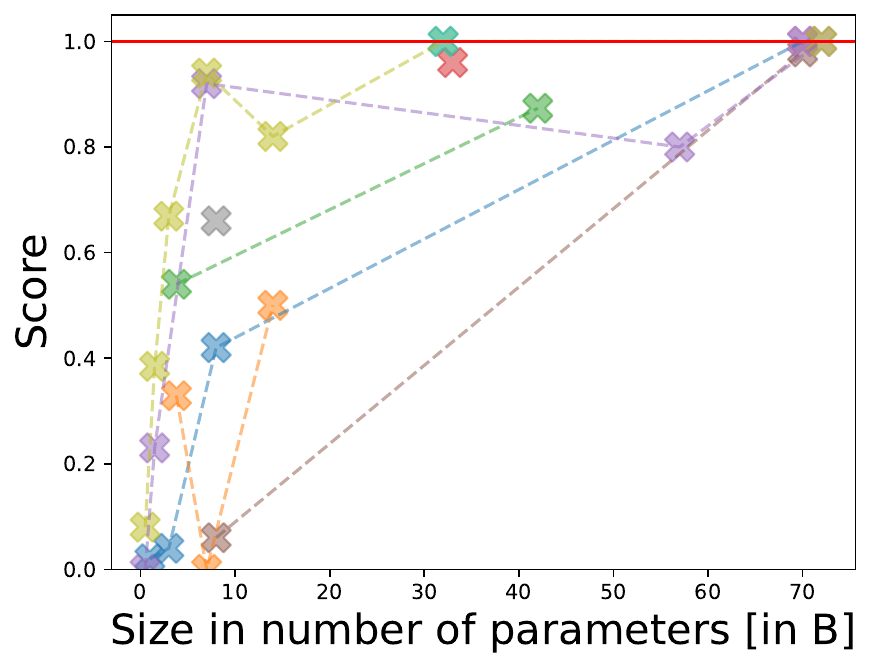}
    \caption{SparqlSyntaxFixing}
    \end{subfigure}\hfill
    \begin{subfigure}{0.33\textwidth}
        \centering
        \includegraphics[width=\linewidth]{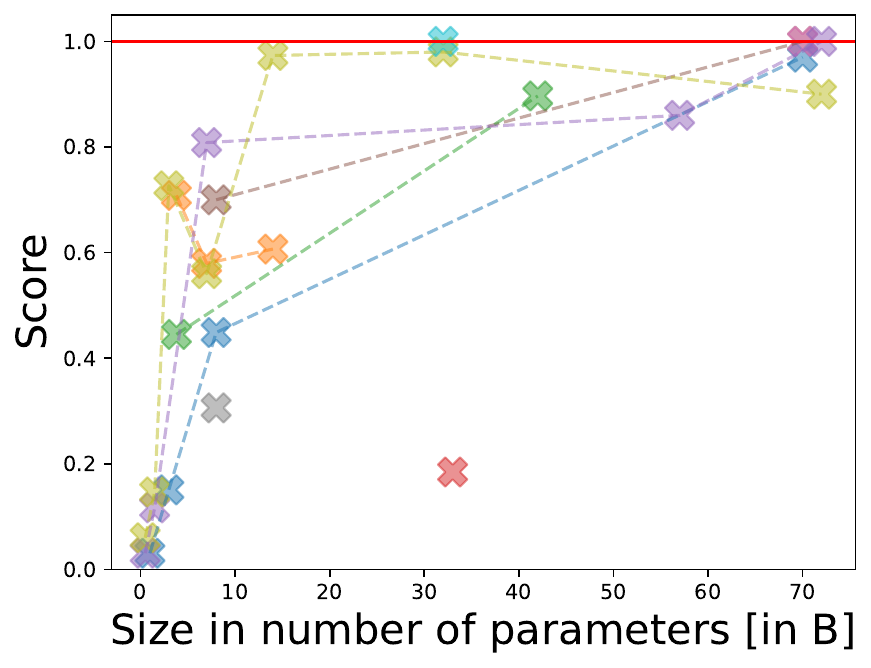}
        \caption{Text2Answer}
    \end{subfigure}
    
    \vskip 0.3cm
    
    \begin{subfigure}{0.33\textwidth}
        \centering
        \includegraphics[width=\linewidth]{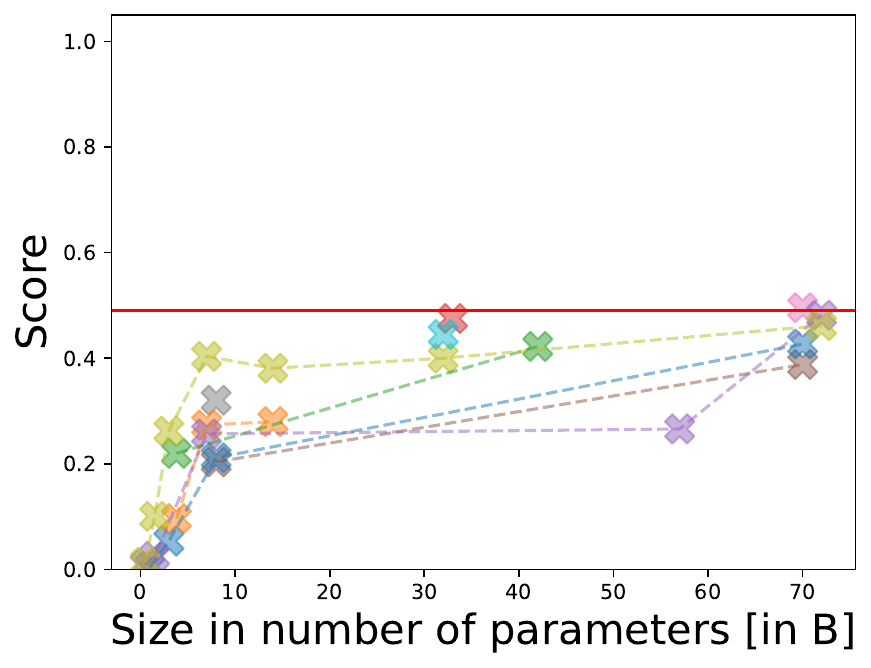}
        \caption{Text2Sparql}
    \end{subfigure}\hfill
    \begin{subfigure}{0.33\textwidth}
        \centering
        \includegraphics[width=0.6\linewidth]{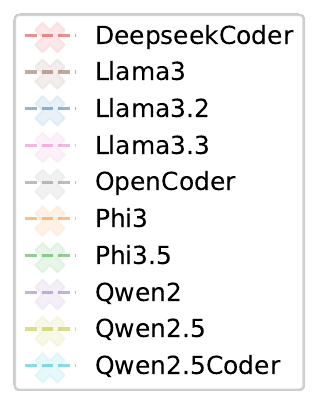}
    \end{subfigure}\hfill
    \vskip 0.3cm
    
    \caption{Plots showing the achieved central measure mean scores per model in relation to the model size for all tasks. Models of the same family are connected through a dashed line. As a reference, for each task, aösp the highest score achieved by a proprietary model included in the benchmark run is given and depicted in a red horizontal line.} 
    \label{fig:3x3_3}
\end{figure}

Complementary to the last section, Figure \ref{fig:3x3_3} shows the average central measure scores for each task class by LLM relative to their sizes. 
In addition, dashed lines connect LLMs of the same family.
For the tasks \textit{Text2Sparql} and \textit{RdfFriendCount}, the plots again expose the overall poor performance of the included LLMs.
Moreover, other previously found patterns are visible in the figures.
Hence, the overall tendency of scores to rise with the model size is noticeable. 
In addition, the plateau and ceiling effects can be seen.

Furthermore, additional insights are visible in the plots.
For reference, the plots also show the highest average score achieved by a proprietary model included in the benchmark run as a red horizontal line.
Here, we see that, except for the \textit{Text2Sparql} task, the best proprietary LLM always reaches an average score of $0.99$ or $1.00$.
With that, the best-performing proprietary model is, except for the \textit{RdfFriendCount} task in which the best open LLMs achieves only a score of $0.55$, on par with the best-performing open LLM. 
Especially the mean scores of tiny $[0-3B]$ and small $(3B-8B]$ models differ greatly.

Moreover, for most tasks, the highest score growth occurs within the range of tiny to smaller medium-sized models around 13B.
Frequently, for tasks showing ceiling effects, already some smaller models of around 8B or 14B reach average scores of $0.8$ or higher.
Here, especially the 8 and 14B Qwen2.5 models stand out.

The two included Mixture-of-Experts (MoE) LLMs, namely Phi-3.5-MoE-instruct (42B parameters thereof 6.6B active) and Qwen2-57B-A14B-Instruct (57B parameters thereof 14B active) show for most tasks scores similar to models of a similar size regarding their total parameters that use all parameters during inference time.
Nevertheless, there are models with lower total parameter counts in the range of the MoE models' active parameter counts that perform comparably.

Besides, the code-specialized models, Qwen2.5-Coder (32B), OpenCoder (8B),
and Deepseek-Coder (33B) are all roughly on par or in a few cases slightly better performing compared to similarly sized LLMs regarding their task performance on the \textit{RdfSyntaxFixing}, \textit{SparqlSyntaxFixing} and \textit{Text2Sparql} that all yield either a RDF graph or a Sparql query. 
Here, DeepSeek-Coder and Qwen2.5-Coder perform similarly to the best-performing open models but this also holds for some non-code specialized models. 
For the other tasks, except for the \textit{RdfFriendCount} task, Qwen2.5-Coder also performs comparably similar or even slightly better than similar-sized models.
In contrast, DeepSeek-Coder and OpenCoder perform worse than other similarly-sized models on the tasks not yielding a KG or SPARQL query.

In the next paragraphs, we look at intra-family and inter-family developments of benchmark scores with respect to the model size.

Models of the same family also reflect the overall tendency of scores to rise with the model size.
In addition, the largest models of the families are typically the best performing.
However, on the family level, occasionally the task performance also drops between size-wise adjacent smaller and larger models.
These drops in performance with larger models typically remain only local and the next larger model of the family often shows a higher or at least steady task performance compared to the model before the drop.
Global, family-overarching saturation effects, i.e., ceiling and plateau effects, are also visible, in particular, for families with many models covering all size bands.

In between different families, in some cases, the score development with rising model size is similar and almost parallel. 
However, in general, clear global parallelisms in the score developments with rising model sizes are not recognizable.

\section{Discussion}
\label{discussion}
In this section, we summarize and discuss key insights from our analysis structured into paragraphs each covering a different insight.

\paragraph{Larger models typically achieve higher scores than smaller ones but there are plateau and ceiling effects.}
In the size category analyses, we saw that most of the time, as expected, typically larger model size groups achieved significantly higher benchmark scores (see also Table \ref{tab:dunnbonferroni}).
For easier tasks, especially the medium and large category pairs got similar high scores.
Hence, here medium-sized models could be a good choice to optimize cost-effectiveness.
In contrast, especially for more difficult tasks, plateaus occurred.
Some were only local and larger models got significantly higher scores than the models within the plateau range.
Consequently, it makes sense to consider the detected local plateaus and decide on a larger LLM since even if the costs increase, also the performance increases significantly.
For global plateaus that are not yet close to the maximum score and spanning up to the large models, it could also make sense to use smaller models since it saves costs and does not affect the task performance much.

\paragraph{Some smaller models also perform comparably well. However, the performance across individual smaller models varies.}
Furthermore, also Figure \ref{fig:3x3_3} confirms that even some small (~8B) or medium-sized (~13B) models might be a good choice since they already achieve reasonably high scores. 
Nevertheless, individual models within the same size band have to be tested explicitly, since their performance also varies. 
However, the insights help to guide the overall model search and indicate whether it seems promising or likely to consider models within a certain size band or not.

\paragraph{Performance drops occur between smaller and larger models from the same family.}
Moreover, within model families, we again saw that local task performance drops can occur between smaller and larger members. 
Hence, it is advisable to also study models adjacent with respect to their size within a family.

\paragraph{The examined Open LLMs cannot cope well with the \textit{RdfFriendCount} and \textit{Text2Sparql} tasks.}
In addition to guiding the model choice within open LLMs, the results also indicate that current state-of-the-art open LLMs as of December 2024 up to a parameter count of 70B, cannot cope well with the \textit{RdfFriendCount} and the \textit{Text2Sparql} tasks. 
Here, the tasks likely require an even larger model.
For the \textit{RdfFriendCount} task, a proprietary LLM included in the benchmark run got a mean central score of $0.99$ or $1.00$.
Hence, here current proprietary models can cope well with the task in contrast to comparably much smaller large open LLMs.
Nevertheless, for the \textit{Text2Sparql} task also the proprietary LLMs achieve not substantially higher mean scores.
Here, the best-performing model achieves only a mean score of $0.49$ indicating that the identified plateau effect even continues for much larger proprietary models.
\paragraph{The examined code-specialized models showed better performance on tasks where a KG or a SPARQL query was requested.}
Compared to the other tasks, especially DeepseekCoder and OpenCoder performed best on the tasks \textit{RdfSyntaxFixing}, \textit{SparqlSyntaxFixing} and \textit{Text2Sparql} in comparison to other tasks. 

\paragraph{The examined Mixture-of-Expert (MoE) models do not show superior performance in comparison with models of the MoE's active parameter count.}
Looking at the individual task scores of models, the MoE LLMs Phi-3.5-MoE-instruct and Qwen2-57B-A14B-Instruct performed mostly comparably to models having a similar total parameter count. However, models of a size in the range of the MoE models' active parameters performed also similarly. 
Hence, for the given tasks, it makes sense to prefer these smaller models instead of the MoE models concerning their cost-effectiveness. 

\section{Conclusion}
\label{conclusion}
In this paper, we analysed the scores of open LLMs from a run of the LLM-KG-Bench benchmark for knowledge graph engineering-related tasks with a focus on the correlation between model size and achieved scores.
Overall, we saw that, as expected, usually the larger the model was, the higher the scores were.
However, our analysis also showed plateau and ceiling effects in which model scores did not differ substantially between smaller and larger models.
Hence, for comparably easy tasks, also smaller models already achieved reasonably high scores.
Consequently, it makes sense to also consider smaller models for similarly complex tasks.
For the \textit{RdfFriendCount} and \textit{Text2Sparql} tasks, the benchmark scores were overall low, plateau effects spanned up to the largest models analysed.
Here, we can conclude that the capabilities of SOTA open LLMs are not yet sufficient to solve tasks of this complexity.
While the \textit{RdfFriendCount} task can be solved by much larger proprietary models, for \textit{Text2Sparql} plateau effects continue, and potentially even larger models are required to sufficiently solve this task.

For future works, we believe that for benchmark runs similar analyses are meaningful to get an overview of the status of SOTA models but also derive generalizable insights that might help to classify whether newly introduced models or models not part of the benchmark run seem promising to consider.
Here, it would be also interesting to examine additionally other scaling law-related factors like the training data, the number of training steps, and factors related to the model architecture.
This would allow for further examination and possible explanations of effects that became apparent in this work like the performance differences between similarly-sized models or performance drops of larger models compared to smaller ones belonging to the same model family. 
Moreover, it is meaningful to extend the LLM-KG Bench framework by more complex variations of already well-solved tasks to be able to figure out whether medium-sized models are still on par with large models in more difficult scenarios. 
Besides, also motivated by the preference of code-specialized LLMs towards tasks requiring a KG or SPARQL query as output, exploring which kind of capabilities are required for specific tasks and why some tasks seem particularly challenging would be a meaningful future contribution to guide targeted solutions.

\begin{acknowledgments}
This work was partially supported by grants from the German Federal Ministry of Education and Research (BMBF) to the projects ScaleTrust (16DTM312D) and KupferDigital2 (13XP5230L) as well as from the German Federal Ministry for Economic Affairs and Climate Action (BMWK) to the KISS project (01MK22001A).
\end{acknowledgments}
\clearpage

\section*{Declaration on Generative AI}
During the preparation of this work, the authors used ChatGPT4o and ChatGPT4.5-RP to: Grammar and spelling check, paraphrase, and reword to improve the writing style.  
After using these tools/services, the authors reviewed and edited the content as needed and take full responsibility for the publication’s content. 

\appendix
\section{Online Resources}
The LLM-KG-Bench framework is available here: \href{https://github.com/AKSW/LLM-KG-Bench}{LLM-KG-Bench}. The raw benchmark run results and further figures are available here: \href{https://github.com/AKSW/LLM-KG-Bench-v3-0-results}{Results LLM-KG-Bench v3}. The code written to perform this analysis can be found here: \href{https://purl.archive.org/llm-kg-bench-run-analysis-code}{Analysis Code}.

\end{document}